%% file: main.tex
\documentclass[lettersize,journal]{IEEEtran}
\usepackage{amsmath,amsfonts}
\usepackage{algorithmic}
\usepackage{algorithm}
\usepackage{array}
\usepackage[caption=false,font=normalsize,labelfont=sf,textfont=sf]{subfig}
\usepackage{textcomp}
\usepackage{stfloats}
\usepackage{url}
\usepackage{verbatim}
\usepackage{graphicx}
\usepackage{cite}
\usepackage{times}
\usepackage{soul}
\usepackage{url}
\usepackage[hidelinks]{hyperref}
\usepackage[utf8]{inputenc}
\usepackage[small]{caption}
\usepackage{graphicx}
\usepackage{amsmath}
\usepackage{amsthm}
\usepackage{booktabs}
\usepackage{algorithm}
\usepackage{algorithmic}
\usepackage[switch]{lineno}
\usepackage{amssymb}
\usepackage{amsfonts}
\usepackage{gensymb}
\usepackage{multirow}
\usepackage[table,xcdraw]{xcolor}
\usepackage{wasysym}

\begin{document}

\title{SiamMo: Siamese Motion-Centric 3D Object Tracking}

\author{Yuxiang Yang, Yingqi Deng, Jing Zhang, Hongjie Gu, Zhekang Dong*~\IEEEmembership{Senior Member,~IEEE,}

\thanks{The project was supported by the National Natural Science Foundation of China (62376080), the Zhejiang Provincial Natural Science Foundation Key Fund of China (LZ23F030003), and the Fundamental Research Funds for the Provincial Universities of Zhejiang (GK239909299001-003).
Corresponding author*: Zhekang Dong (englishp@126.com)
}
\thanks{Y. Yang, Y. Deng, H. Gu, and Z. Dong are with the School of Electronics and Information, Hangzhou Dianzi University, Hangzhou 310018, China. J. Zhang is with the School of Computer Science, The University of Sydney, NSW 2006, Australia.} 
}

\maketitle

\begin{abstract}

Current 3D single object tracking methods primarily rely on the Siamese matching-based paradigm, which struggles with textureless and incomplete LiDAR point clouds. Conversely, the motion-centric paradigm avoids appearance matching, thus overcoming these issues. However, its complex multi-stage pipeline and the limited temporal modeling capability of a single-stream architecture constrain its potential. In this paper, we introduce SiamMo, a novel and simple Siamese motion-centric tracking approach. Unlike the traditional single-stream architecture, we employ Siamese feature extraction for motion-centric tracking. This decouples feature extraction from temporal fusion, significantly enhancing tracking performance. Additionally, we design a Spatio-Temporal Feature Aggregation module to integrate Siamese features at multiple scales, capturing motion information effectively. We also introduce a Box-aware Feature Encoding module to encode object size priors into motion estimation. SiamMo is a purely motion-centric tracker that eliminates the need for additional processes like segmentation and box refinement. Without whistles and bells, SiamMo not only surpasses state-of-the-art methods across multiple benchmarks but also demonstrates exceptional robustness in challenging scenarios. SiamMo sets a new record on the KITTI tracking benchmark with 90.1\% precision while maintaining a high inference speed of 108 FPS. The code will be released at \href{https://github.com/HDU-VRLab/SiamMo}{SiamMo}.

\end{abstract}

\begin{IEEEkeywords}
3D Object Tracking, Point Cloud, Siamese network, Convolutional Neural Networks
\end{IEEEkeywords}

\section{Introduction}

\begin{figure}[t]
    \centering
    \includegraphics[width=\linewidth]{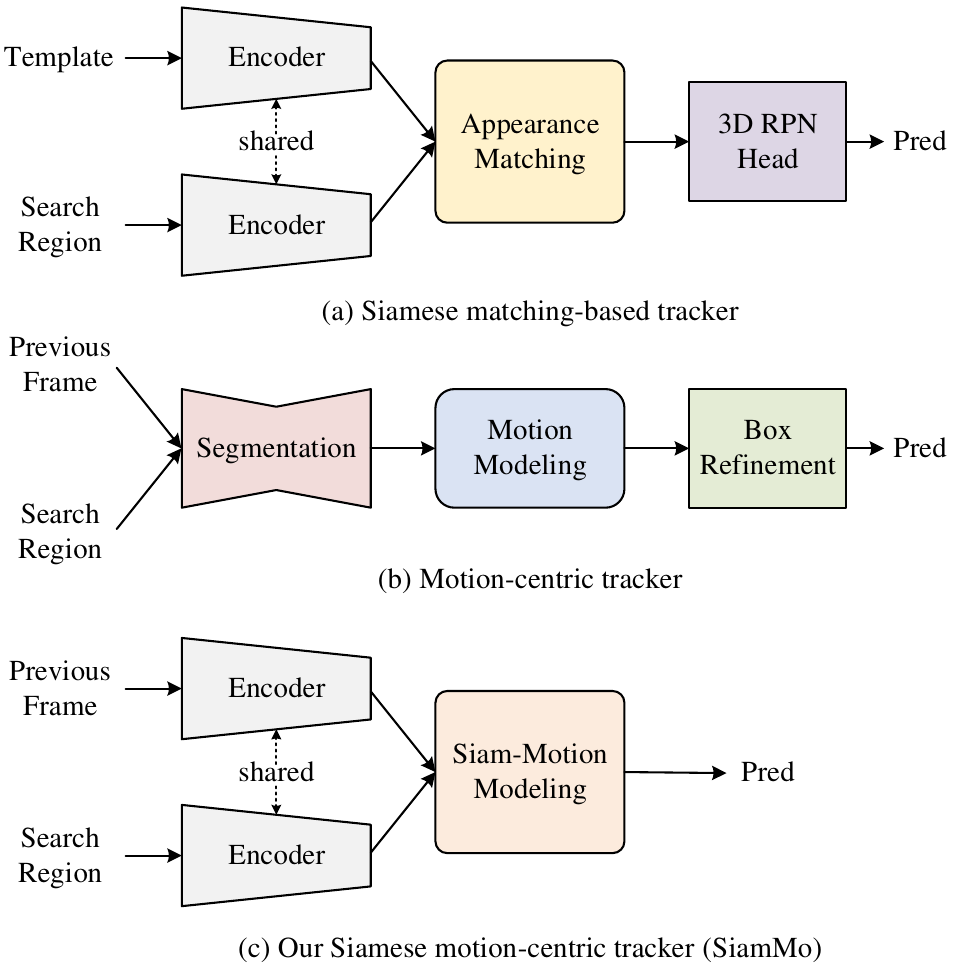}
    \caption{Comparison of typical 3D object tracking methods. (a) Siamese matching-based tracker that relies on appearance matching. (b) Motion-centric tracker that requires multiple stages of segmentation and box refinement. (c) Our Siamese motion-centric tracker adopts Siamese architecture to perform motion-centric tracking in an end-to-end simple single-stage manner.} 
    \vspace{-5pt}
    \label{fig:paradigm}
\end{figure}

3D single object tracking (SOT) in point clouds is a fundamental task in computer vision. It aims to localize a specific target across a sequence of point clouds, given only its initial status. In recent years, the advancement of 3D data acquisition devices has sparked growing interest in utilizing point clouds for a range of vision tasks, including object detection~\cite{voxel-transformer,group-free,tcsvt4,tcsvt5} and tracking~\cite{m2trackvanilla,p2b}. Point cloud-based object tracking, in particular, has seen significant advancements, driven by its immense potential in applications like surveillance, autonomous driving. Nonetheless, it remains challenging due to the substantial appearance variations of targets and the inherent sparsity of 3D point clouds resulting from occlusion and limited sensor resolution.

Existing matching-based SOT methods~\cite{p2b,bat,v2b,pttr,stnet} generally utilize certain forms of Siamese networks for feature extraction, as shown in Fig.~\ref{fig:paradigm}(a). After transforming the cropped target template and search embeddings in the same feature space with a shared encoder, these methods enhance target-specific features with various appearance matching techniques, such as cosine similarity~\cite{p2b,bat,v2b} and cross-attention~\cite{pttr,stnet}. Although the Siamese matching-based paradigm has become a popular design in existing models, appearance matching has long suffered from issues with textureless and incomplete LiDAR point clouds. Beyond this paradigm, as shown in Fig.~\ref{fig:paradigm}(b), a new motion-centric tracker M$^2$-Track~\cite{m2track} offers a new perspective for 3D SOT. It takes as input point clouds without cropping from two successive frames, explicitly modeling the relative target motion in a single-stream architecture, largely overcoming the challenges. However, it fuses adjacent point clouds and processes them in a single-stream architecture, lacking explicit target information from adjacent frames for accurate localization. To compensate for this deficiency, M$^2$-Track requires additional segmentation and box refinement, which makes the training objective complex and results in cumulative errors.

In this paper, we propose a novel Siamese motion-centric tracking approach, dubbed SiamMo, as shown in Fig.~\ref{fig:paradigm}(c). It adopts a simple single-stage tracking pipeline of Siamese feature extraction and motion modeling. To learn better features for point clouds of varying density, we first divide non-uniform points into regular voxels and then adopt two weight-sharing encoders that embed voxelized point clouds of successive frames into the same feature space. This decouples feature extraction from temporal fusion, significantly enhancing tracking performance. Subsequently, we design a Spatio-Temporal Feature Aggregation (STFA) module that integrates the encoded features at multiple scales, thus effectively capturing the motion information. Moreover, we introduce a Box-aware Feature Encoding (BFE) strategy that injects explicit box priors into motion features for prediction. Despite being conceptually simple, our BFE can boost tracking performance, with negligible computation. In a nutshell, using neither segmentation nor box refinement, SiamMo achieves precise localization by directly inferring the relative target motion in a single-stage manner. Experiments demonstrate our SiamMo method is effective, achieving new state-of-the-art performance on three challenging tracking benchmarks, \textit{i.e.}, KITTI~\cite{kitti}, NuScenes~\cite{nuscenes}, and Waymo Open Dataset~\cite{wod}. For example, SiamMo pushes the current best KITTI tracking result to a new record with 90.1\% precision, while maintaining a high inference speed of 108 FPS. On the challenging NuScenes dataset, SiamMo outperforms the previous motion-centric tracker M$^2$-Track by 11.08\% in success rate. Moreover, SiamMo shows excellent robustness under the sparse settings and scenarios with distractors.

In summary, our main contributions are as follows:
\begin{itemize}
\item We propose a novel Siamese motion-centric tracking approach SiamMo, which adopts Siamese architecture and models target motion in a simple single-stage pipeline.
\item We design the STFA module that integrates features from the adjacent frames at multiple scales to effectively capture the motion information. Furthermore, we design the BFE module that provides target box priors for motion prediction in a simple way.
\item SiamMo surpasses state-of-the-art methods on three challenging benchmarks while demonstrating excellent robustness and maintaining a high inference speed.
\end{itemize}

\section{Related Work}
\subsection{3D Single Object Tracking}
Existing methods for 3D single object tracking in point clouds predominantly follow the Siamese matching-based paradigm, utilizing Siamese networks for feature extraction and appearance matching. Siamese networks are general models for comparing entities, which have various applications including face verification~\cite{siamese-face}, visual tracking~\cite{siamrpn++}, one-shot learning~\cite{siam-one-shot}, and unsupervised learning~\cite{simsiam}. In 3D Siamese matching-based object tracking~\cite{sc3d,p2b}, the inputs to Siamese networks are point clouds from the cropped target template and search region, and the matching is conducted to mine resembling patterns of the target and reveal the local tracking clue.

As a pioneer in the realm of 3D single object tracking in point clouds, SC3D~\cite{sc3d} is the first 3D Siamese tracker. It exhaustively compares the target template with numerous candidate patches generated from the current frame. However, this results in multiple forward passes of the model, leading to significant computation overhead. To address this issue, P2B~\cite{p2b} introduces an end-to-end Siamese region proposal network. Initially, it employs a Siamese backbone to encode both the template and the search region point clouds. Subsequently, it enhances target-specific features through appearance matching. Finally, P2B integrates VoteNet~\cite{votenet} to generate 3D proposals and selects the proposal with the highest score as the target. This approach significantly enhances tracking performance while maintaining real-time speed. Therefore many follow-up works adopt the same Siamese paradigm. BAT~\cite{bat} designs a box-aware feature fusion module to enhance target-specific features. OSP2B~\cite{osp2b} introduces a one-stage point-to-box network to improve the 3D Siamese paradigm. Inspired by the success of transformers~\cite{transformer}, PTT~\cite{ptt}, PTTR~\cite{pttr}, STNet~\cite{stnet} and GLT-T~\cite{glt} develop various attention mechanisms to model long-range dependency for object tracking. CXTrack~\cite{cxtrack} exploits contextual information across consecutive frames to improve tracking accuracy. TAT~\cite{tat} proposes a temporal-aware Siamese tracking method that aggregates target information in multiple historical templates. SyncTrack~\cite{synctrack} unifies the process of feature extraction and target information integration. 

Beyond 3D Siamese matching-based tracking, M$^2$-Track~\cite{m2trackvanilla} takes point clouds without cropping from two frames as input, and then segments the target points from their surroundings. After that, M$^2$-Track explicitly models the relative target motion with the cropped points to coarsely localize the target and finally refines the result through a motion-assisted shape completion strategy. In contrast, we adopt a Siamese architecture and directly model the target motion in a simple single-stage pipeline, without using either segmentation or box refinement.

\subsection{3D Multi-Object Tracking}
3D multi-object tracking models tracklets of multiple objects along multi-frame LiDAR. The majority of prior works in 3D MOT leverage the LiDAR modality. Due to the recent advances in LiDAR-based 3D detection~\cite{voxelnext,tcsvt1,tcsvt2,tcsvt3}, especially the reliable range information, most state-of-the-art 3D MOT algorithms adopt a “tracking-by-detection” paradigm. Objects are first detected in each frame, followed by heuristic association of detected bounding boxes based on object motion or appearance~\cite{pnpnet,pc3t}. A recent development, SimTrack~\cite{simtrack} proposes a unified pipeline that combines object detection and motion association for joint detection and tracking. Moreover, there is a growing trend in leveraging both LiDAR and camera data for 3D MOT~\cite{bevfusion,camo-mot,eagermot}. These multi-sensor fusion techniques enhance robustness against occlusion and object misdetection by combining data from multiple sensors. Our motion-centric tracker is inspired by MOT's motion-based association. However, unlike MOT, which relies on motion estimation from detection results, our approach is detector-independent and learns fine-grained target-specific features.

\subsection{Spatio-temporal Learning on Point Clouds}
Our method employs spatio-temporal learning to infer relative motion across multiple frames. In recent years, two primary approaches have emerged for analyzing 3D dynamic point cloud sequences. Voxelization-based methods, such as MinkowskiNet~\cite{MinkowskiNet}, convert sequences into 4D occupancy grids and utilize sparse convolution for processing. Another technique, 3D Sparse Conv LSTM~\cite{lstm}, voxelizes sequences into sparse 4D voxels, convolving them on a sparse grid within an LSTM network to capture temporal dependencies. Some methods operate directly on raw point cloud sequences. MeteorNet~\cite{meteornet}, inspired by PointNet~\cite{pointnet,pointnet++}, employs grouping techniques to construct spatio-temporal neighborhoods and aggregate inter-neighborhood features. ASAP-Net~\cite{asap} utilizes a novel spatio-temporal correlation grouping approach and an attentive temporal embedding layer to recurrently fuse inter-frame local features. Our SiamMo integrates features from the adjacent frames at multiple scales to effectively capture the motion information.

\begin{figure*}[tp]
    \includegraphics[width=\textwidth]{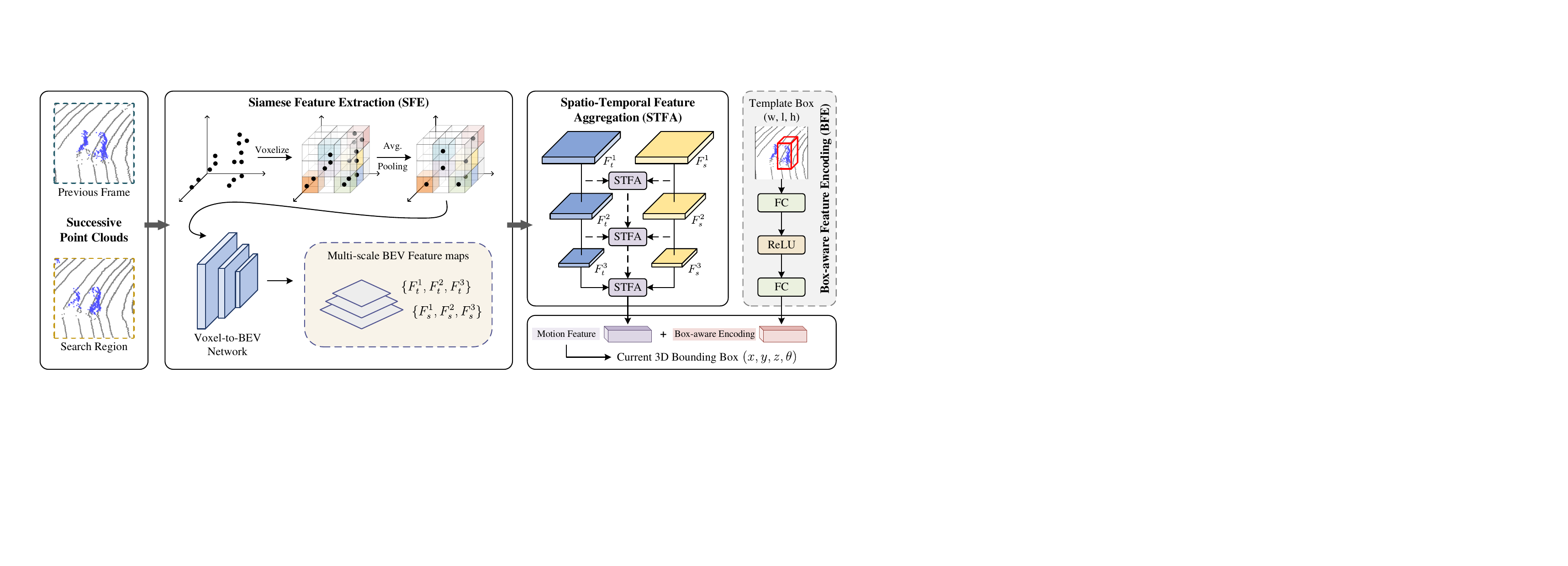}
    \centering
    \caption{Architecture of SiamMo. SiamMo comprises three main blocks. SFE encodes two successive frames into multi-scale BEV feature maps. STFA then integrates the BEV feature maps at multiple scales. Finally, BFE injects the box priors into motion features for prediction.}
    \label{fig:SiamMo}
\end{figure*}

\section{Method}

\subsection{Overview}
Given an initial target bounding box (BBox) $\mathcal{B}_1$ and a sequence of point clouds $\{\mathcal{P}_t\}_{t=1}^{T}$, the objective of 3D SOT is to localize a series of target BBoxes $\{\mathcal{B}_t=(x_t,y_t,z_t,w_t,l_t,h_t)\}_{t=2}^{T}$ in subsequent frames, where $T$ represents the number of frames. The parameters $(x,y,z)$, $(w,l,h)$, and $\theta$ define the BBox center, size, and rotation angle around $up$-axis, respectively. At each timestamp $t$, the LiDAR point clouds $\mathcal{P}_t \in \mathbb{R}^{N_t\times 3}$ contain $N_t$ points with 3 channels encoding the $xyz$ global coordinates. Given that the size of the tracking target is assumed to have minimal variation across frames, we focus primarily on regressing the parameters $(x_t,y_t,z_t,\theta_t)$ for $\mathcal{B}_t$.

At timestamp $t$ ($t>1$), the point clouds of two consecutive frames $\mathcal{P}_{t-1}$ and $\mathcal{P}_t$ as well as the previously predicted target BBox $\mathcal{B}_{t-1}$ in $\mathcal{P}_{t-1}$ are known. We first transform the point clouds from the world coordinate system to the canonical system w.r.t the $\mathcal{B}_{t-1}$. We predict the relative target motion (RTM) between the two successive frames. RTM is a rigid-body transformation that is defined between two target bounding boxes. Since objects of interest are always aligned with the ground, we only consider 4DOF RTM, which comprises a translation offset $(\Delta x_t,\Delta y_t,\Delta z_t)$ and a yaw offset $\Delta \theta_t$. By applying the RTM $\mathcal{M}_{t-1,t}$ to the BBox $\mathcal{B}_{t-1}$, the current BBox $\mathcal{B}_t$ can be obtained. The prediction process can be formulated as:
\begin{equation}
    \mathcal{F}(\mathcal{P}_{t-1},\mathcal{P}_t,\mathcal{B}_{t-1}) \mapsto (\Delta x_t,\Delta y_t,\Delta z_t,\Delta \theta_t).
\label{definition}
\end{equation}

Based on Eq.~\eqref{definition}, we propose a novel approach named SiamMo for Siamese motion-centric tracking. As shown in Fig.~\ref{fig:SiamMo}, SiamMo consists of three main parts: 1) \textit{Siamese Feature Extraction} (SFE), 2) \textit{Spatio-Temporal Feature Aggregation} (STFA), and 3) \textit{Box-aware Feature Encoding} (BFE). SFE first divides the raw point clouds into regular voxels, and then encodes them to multi-scale bird's eye view (BEV) feature maps with Voxel-to-BEV Network. Next, the BEV feature maps are aggregated with STFA, fusing supplementary information from the adjacent frames at multiple scales. For rigid objects, BFE will inject target box priors into motion features. Finally, a simple multi-layer perceptron (MLP) is used for motion prediction.

\subsection{Siamese Feature Extraction}
\paragraph{Voxelization} LiDAR-scanned point clouds exhibit a varying sparsity property --- points in close proximity are densely packed, whereas those farther away are significantly sparser. Learning point features that effectively accommodate both densely and sparsely populated regions present a considerable challenge. To mitigate this issue, we propose to utilize the voxel-based representation~\cite{voxelnet} for feature extraction, rather than point-based representations~\cite{pointnet,pointnet++,dgcnn} as in previous works. Specifically, we voxelize the two successive point clouds $\mathcal{P}_{t-1}^{'}$ and $\mathcal{P}_{t}^{'}$ into regular voxels respectively. The initial feature of each non-empty voxel is calculated as the mean values of point coordinates within the voxel:
\begin{equation}
    v_k = \frac{1}{N}\sum_{i=1}^{N}p_k(i),
\label{voxelization}
\end{equation}
where $v_k$ represents the $k$th element of a non-empty voxel that contains $N$ points. $p_k(i)$ is the $i$th point with the 3D coordinates $(x, y, z)$ in the $k$th voxel. As illustrated in Fig.~\ref{fig:SiamMo}, 
voxel features exhibit reduced sensitivity to point count variations, thus mitigating the varying sparsity issue to some extent.

\paragraph{Voxel-to-BEV Network} To learn discriminative features for both the previous frame and search region, we present a top-down convolutional network based on Siamese architecture. The network consists of two parts: \textit{sparse voxel feature extraction} (SVFE), and \textit{dense BEV feature extraction} (DBFE). In each layer of SVFE, we stack two submanifold convolution layers~\cite{submanifold} with kernel size 3 $\times$ 3 $\times$ 3 and stride 1. Between the layers, we insert a 3D sparse convolution block~\cite{sparseconv} with kernel size 3 $\times$ 3 $\times$ 3 and stride 2 to downsample the spatial resolution. After four layers of 8 times down-sampling, the spatial resolution becomes increasingly small as well as the features are dense and rich in information. To further alleviate the adverse effect of varying sparsity, we flatten the sparse 3D features along the Z-axis to obtain dense BEV features. By stacking a series of 2D standard convolution layers with kernel size 3 $\times$ 3, DBFE produces multi-scale BEV feature maps with the feature strides of $\{1,2,4\}$. The multi-scale BEV feature maps of the previous frame and search region are denoted as $\{F_t^1,F_t^2,F_t^3\}$ and $\{F_s^1,F_s^2,F_s^3\}$, respectively.

\subsection{Spatio-Temporal Feature Aggregation}
STFA aims to integrate features from consecutive frames at multiple scales for motion modeling. 

Intuitively, effective motion modeling necessitates rich representations at multiple scales. Integrating these multifaceted representations is expected to significantly enhance the network's capability to accurately localize the targets with various motion patterns.

\begin{figure}[tp]
    \centering
    \includegraphics[width=\linewidth]{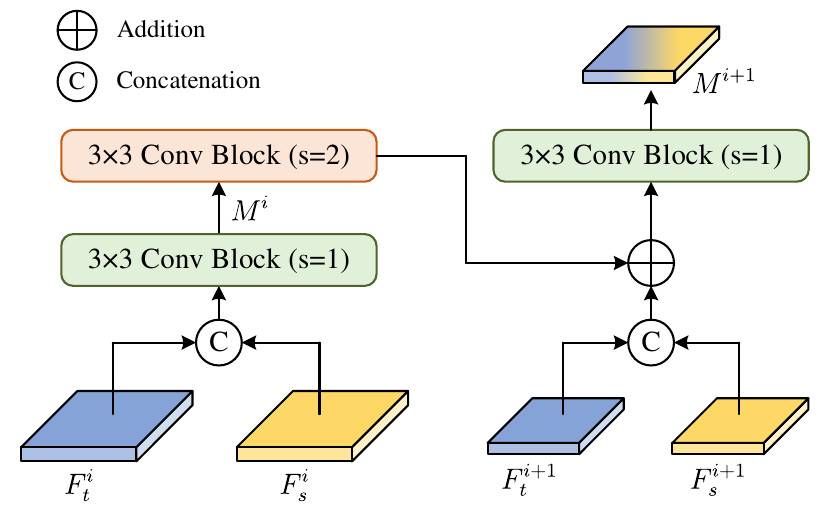}
    \caption{Diagram of Spatio-Temporal Feature Aggregation.} 
    \label{fig:STFA}
\end{figure}

As shown in Fig.~\ref{fig:STFA}, STFA integrates multi-scale BEV feature maps $\{F_t^1,F_t^2,F_t^3\}$ and $\{F_s^1,F_s^2,F_s^3\}$ to propagate the target information from the previous frame into the search region and extract motion features at different scales. At each scale, the motion feature $M^i$ is obtained as:

\begin{equation}
M^1=Conv_{3\times3}([F_t^{1},F_s^{1}]),
\label{eq:m1}
\end{equation}
\begin{equation}
M^{i+1}=Conv_{3\times3}(Down(M^i)+[F_t^{i+1},F_s^{i+1}]),
\end{equation}

where $i \in \{1,2\}$ denotes scale index, $[\cdot,\cdot]$ denotes channel-wise concatenation, $Down$ denotes down-sampling, and $Conv_{3\times3}$ denotes a convolutional layer with a 3 $\times$ 3 kernel and a stride of 1. In the first layer, $M^1$ is obtained by Eq.~\ref{eq:m1}. In the subsequent layers, the feature $M^i$ is resized to the same size as $F_c^{i+1}$ at two different scales via down-sampling. Then, they are summed element-wise followed by a convolutional layer to produce the motion feature $M^{i+1}$ at the next scale. In the last layer, a global max-pooling (GMP) is applied on top of the final motion feature to derive the output motion feature $M^{out}$.

\subsection{Box-aware Feature Encoding}
As shown in previous works~\cite{bat,m2track}, the initial frame's ground truth bounding box serves as a strong target-specific cue, thereby substantially enhancing tracking performance. BAT~\cite{bat} improves the feature comparison by designing a size-aware and part-aware BoxCloud feature. M$^2$-Track~\cite{m2track} extends the point coordinates to the box-aware features to improve the target segmentation and localization accuracy.

In contrast to these works, we propose the BFE inspired by positional encoding~\cite{transformer}. Specifically, BFE first encodes the template box size $S$ that contains three parameters $(w,l,h)$ to the box-aware encoding $E_{box}$ with an MLP. $E_{box}$ has the same channel dimension as the output motion feature $M^{out}$, and the MLP contains two fully-connected layers and a ReLU activation layer~\cite{relu}. Then BFE adds the box-aware encoding $E_{box}$ and the output motion feature $M^{out}$. Finally, the added feature is fed into an MLP to regress the RTM $\mathcal{M}_{t-1,t}$. In this way, we can disentangle the box-aware feature encoding from the point cloud feature extraction, enhancing the tracking robustness. The process could be formulated as follows:
\begin{equation}
    \mathcal{M}_{t-1,t}=MLP^{'}(M^{out}+MLP(S)).
\end{equation}

\input{Table/kitti}

\subsection{Implementation Details}
\paragraph{Training \& Inference} The training objective is to minimize a single regression loss for motion modeling. There are many choices of loss functions, learned and fixed. In this work, we adopt residual log-likelihood estimation~\cite{rle} to achieve accurate regression for targets with diverse attributes (\textit{e.g}., sizes, and motion patterns). We implement our network in PyTorch~\cite{pytorch} using the open-sourced MMEngine~\cite{mmengine}. We train the network end-to-end on 2 NVIDIA GTX 4090 GPUs for 400 epochs using the AdamW~\cite{adamw} optimizer with a learning rate of 1e-4. We use a batch size of 256 and a weight decay of 0.01. During the inference, the model tracks a target frame-by-frame in a point cloud sequence given the target BBox at the first frame.

\paragraph{Network Details} For voxelization, we use a point cloud range of [(-4.8, 4.8), (-4.8, 4.8), (-1.5, 1.5)] meters and a voxel size of [0.075, 0.075, 0.15] meters for \textit{Car} and \textit{Van}. For non-rigid objects including \textit{Pedestrian} and \textit{Cyclist}, we adopt a range of [(-1.92, 1.92), (-1.92, 1.92), (-1.5, 1.5)] meters and a voxel size of [0.03, 0.03, 0.15] meters. For large objects such as \textit{Truck}, \textit{Trailer}, and \textit{Bus}, we adopt a range of [(-9.6, 9.6), (-9.6, 9.6), (-3.0, 3.0)] meters and a voxel size of [0.15, 0.15, 0.3] meters. The spatial feature volumes in SVFE have four scales with feature dimensions of 16, 32, 64, and 128. There are three blocks in DBFE with a stride of $\{1,2,4\}$ and feature dimensions of 128, 256, and 256.

\paragraph{Augmentation} To simulate inaccurate predictions, we randomly rotate the target BBox at $(t-1)$ timestamp around its vertical axis by \textit{Uniform} [-6\degree, 6\degree], and translate it by offsets drawn from \textit{Uniform} [-0.3, 0.3] meter during training. We denote the perturbed BBox as $\mathcal{B}_{ref}$. Then, we transform the point clouds and the GT BBox at $t$ timestamp from the world coordinate system to the canonical box coordinate system w.r.t the $\mathcal{B}_{ref}$. To encourage the model to learn various motions during training, we randomly translate the point clouds and the BBox at $t$ timestamp by offsets drawn from a Gaussian distribution $\mathcal{N}(0,\Sigma)$, where $\Sigma=(0.3,0.2,0.1)$ w.r.t 3D coordinates $(x,y,z)$ respectively. To further increase data diversity and enhance model robustness, we augment the training data by randomly flipping it across both spatial and temporal dimensions. Concretely, we flip both point clouds and BBoxes horizontally and point cloud sequences along the temporal dimension. 

\section{Experiments}
\subsection{Experimental Settings}
\paragraph{Datasets} We validate the effectiveness of SiamMo on three widely-used challenging datasets: KITTI~\cite{kitti}, NuScenes~\cite{nuscenes}, and Waymo Open Dataset (WOD)~\cite{wod}. KITTI contains 21 video sequences for training and 29 video sequences for testing. We follow previous work~\cite{p2b} to split the training set into train/val/test splits due to the inaccessibility of the labels of the test set. NuScenes contains 1,000 scenes, which are divided into 700/150/150 scenes for train/val/test. Following the implementation in~\cite{bat}, we compare with the previous methods on five categories including \textit{Car}, \textit{Pedestrian}, \textit{Truck}, \textit{Trailer}, and \textit{Bus}. For WOD, we follow LiDAR-SOT~\cite{lidar-sot} to evaluate our method, dividing it into three splits, \textit{i.e.}, easy, medium, and hard, according to the number of points in the first frame of each track. Following STNet~\cite{stnet}, we use the trained model on the KITTI dataset for evaluation on the WOD to assess the generalization ability of our 3D tracker.

\paragraph{Evaluation Metrics} Following~\cite{p2b}, we adopt Success and Precision defined in one pass evaluation (OPE)~\cite{metric} as the evaluation metrics. Success denotes the Area Under Curve (AUC) for the plot showing the ratio of frames where the Intersection Over Union (IoU) between the predicted box and the ground truth is larger than a threshold, ranging from 0 to 1, while Precision denotes the AUC for the plot showing the ratio of frames where the distance between their centers is within a threshold, ranging from 0 to 2 meters.

\input{Table/waymo}

\input{Table/nuscenes}

\subsection{Comparison with State-of-the-art Methods}
\begin{figure*}[t]
    \includegraphics[width=1\textwidth]{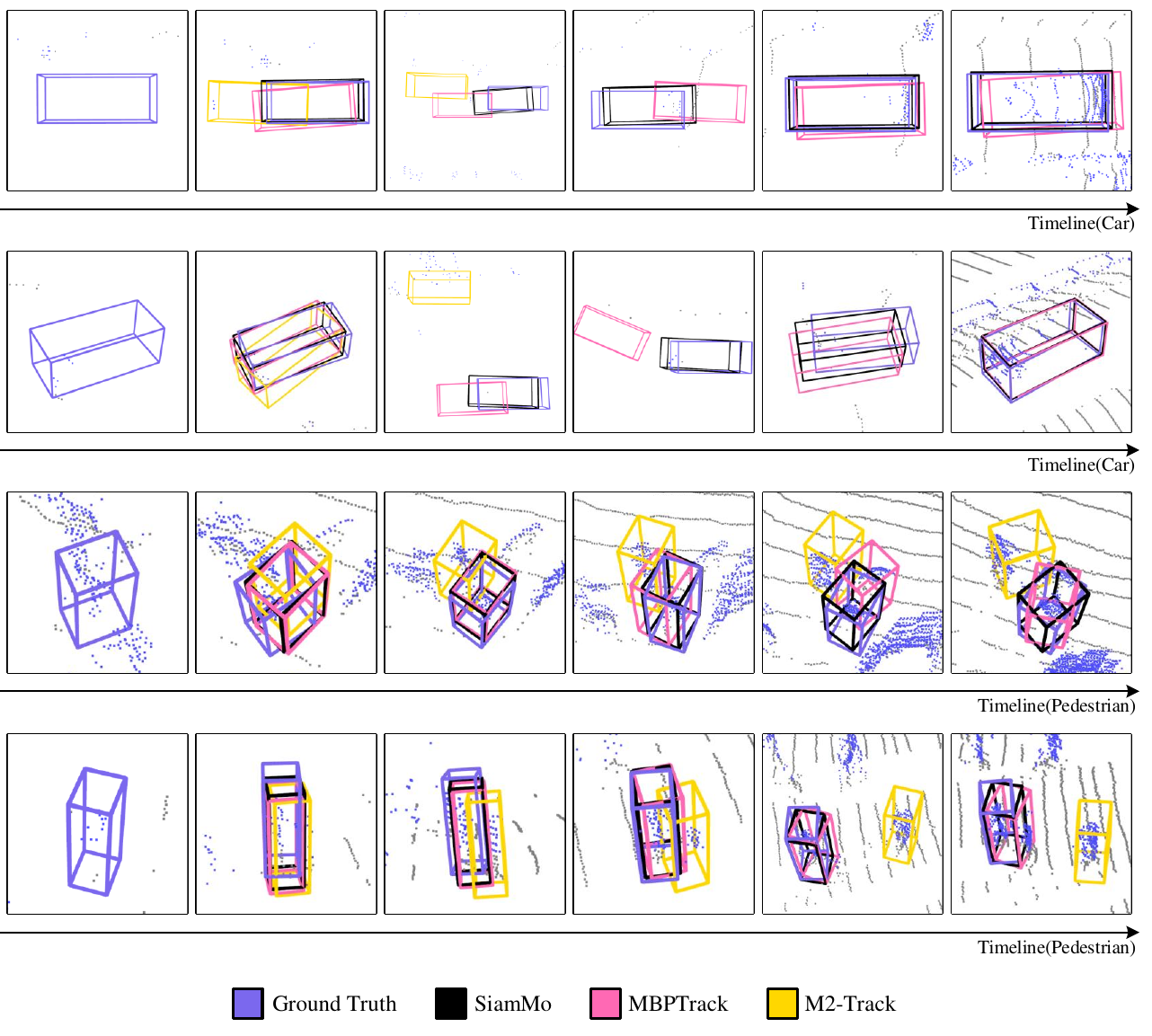}
    \centering
    \caption{Visualization of tracking results by our SiamMo and state-of-the-art methods.}
    \label{fig:vis}
\end{figure*}

\paragraph{Results on KITTI \& WOD} We present a comprehensive comparison on the KITTI dataset between the proposed method and previous state-of-the-art methods. As shown in Table~\ref{kitti}, SiamMo exhibits superior performance across various categories, achieving the highest mean Success and Precision rates of 72.3\% and 90.1\%, respectively. Moreover, SiamMo performs better than the recent state-of-the-art method MBPTrack~\cite{mbptrack} by a gain of 2.0\% and 2.2\% in the mean Success and Precision rates, respectively. Compared to the motion-centric tracker M$^2$-Track~\cite{m2track}, SiamMo unleashes the potential of motion-centric tracking by a significant margin of 9.4\% and 6.7\% in terms of mean Success and Precision, respectively. We also visualize the tracking results for qualitative comparisons in Fig.~\ref{fig:vis}. In the Car sequence where the point clouds are extremely sparse, M$^2$-Track loses the target during tracking. Both SiamMo and MBPTrack keep the correct tracking to the last frame, while SiamMo shows more accurate localization. In the Pedestrian sequence, M$^2$-Track chooses the wrong objects to track when intra-class distractors are close, while SiamMo is able to generate stable and reliable predictions.

To validate the generalization ability of the proposed methods, we follow previous methods~\cite{v2b,stnet} by evaluating the KITTI pre-trained Car and Pedestrian models on the WOD dataset~\cite{wod}. As shown in Table~\ref{waymo}, our SiamMo outperforms other comparison methods under different levels of sparsity, especially in the Pedestrian category. In summary, our proposed approach accurately tracks targets within challenging scenes and exhibits strong generalization capabilities across unseen scenarios.

\paragraph{Results on NuScenes} NuScenes~\cite{nuscenes} poses a more formidable challenge for the 3D SOT task compared to KITTI~\cite{kitti}, primarily attributable to its larger data volumes and sparser annotations (2Hz for NuScenes versus 10Hz for KITTI and Waymo Open Dataset). Following the methodology established by M$^2$-Track~\cite{m2track}, we perform a comparative evaluation on the NuScenes dataset against prior methods, including SC3D~\cite{sc3d}, P2B~\cite{p2b}, PTT~\cite{ptt}, BAT~\cite{bat}, PTTR~\cite{pttr}, GLT-T~\cite{glt}, MBPTrack~\cite{mbptrack} and M$^2$-Track. As shown in Table~\ref{nuscenes}, SiamMo exceeds all the competitors under all categories, mostly by a large margin. On this large-scale dataset, SiamMo surpasses M$^2$-Track more pronounced (11.08\% success and 9.95\% precision gains).

\begin{figure}[t]
    \centering
    \includegraphics[width=\linewidth]{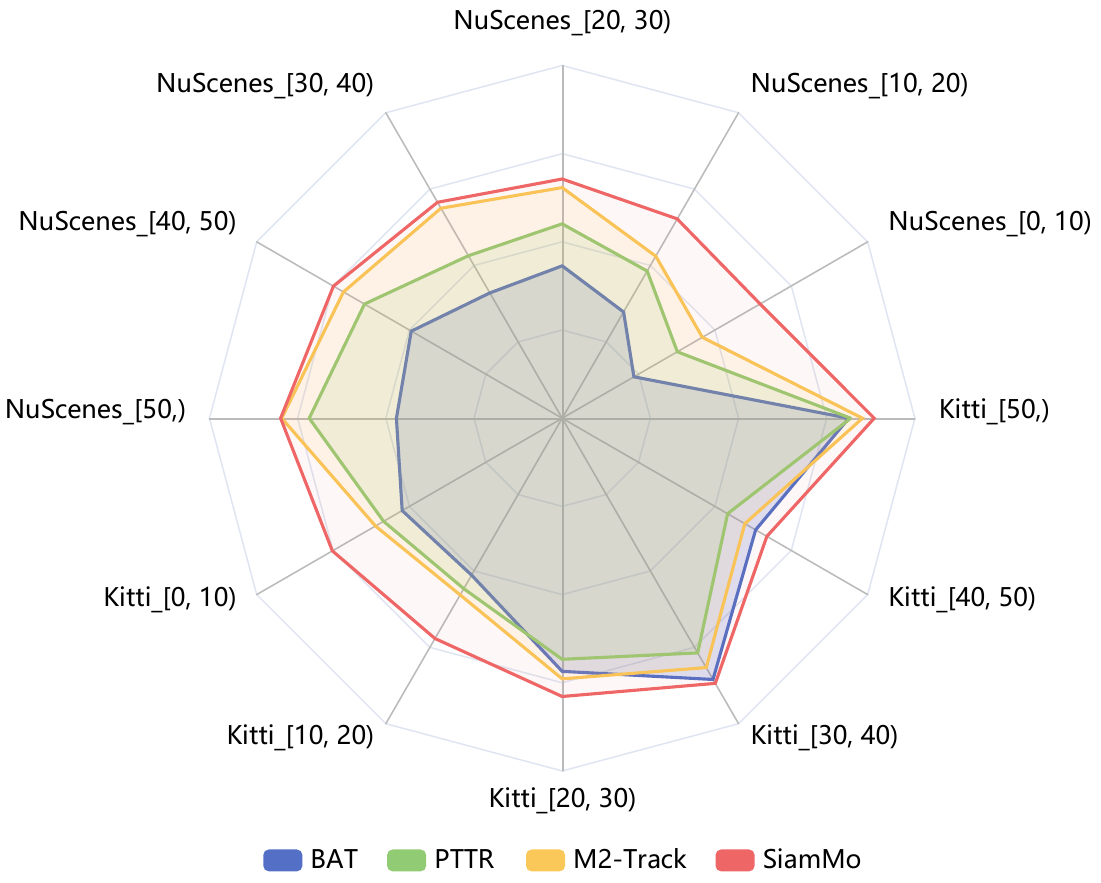}
    \caption{Robustness to sparsity. [a, b) is the number of points in the first frame's car. We use the Success as the evaluation metric. The range of the number line is from 0.2 to 0.9.}  
    \label{fig:sparsity}
\end{figure}

\begin{figure}[b]
    \centering
    \includegraphics[width=\linewidth]{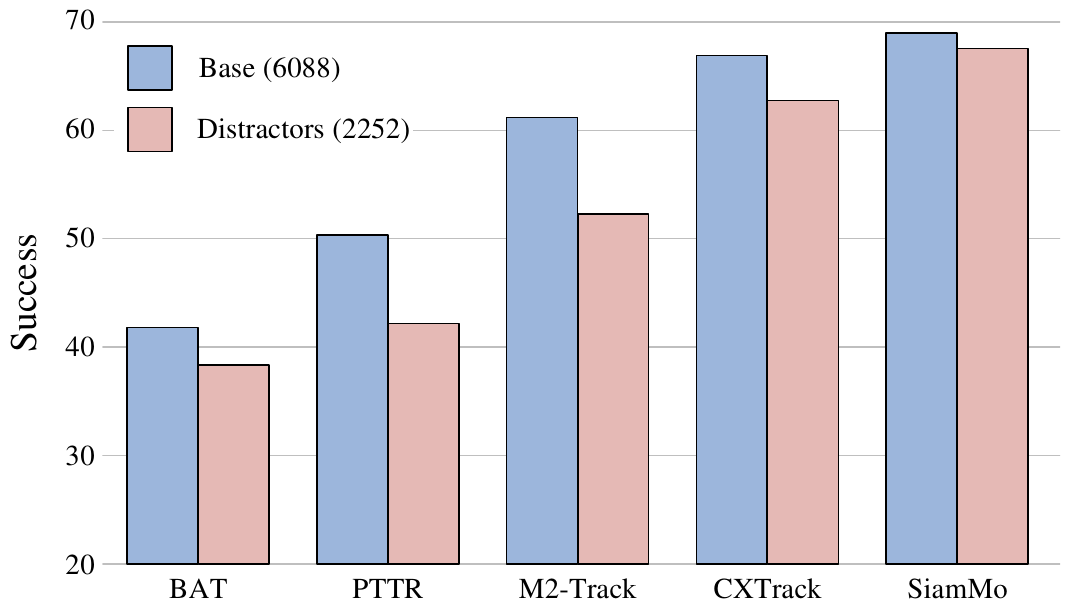}
    \caption{Robustness to distractors. We use the Success as the evaluation metric. Please note the performance drop between the bar [$\color[RGB]{156,182,220}{\blacksquare}$] and the bar [$\color[RGB]{229,185,181}{\blacksquare}$].}  
    \label{fig:distractors}
\end{figure}

\input{Table/siamese}
\subsection{Ablation Study and Analysis}
\paragraph{Effectiveness of Siamese architecture} We investigate the effectiveness of the Siamese architecture in Table~\ref{siamese}. For experiments that utilize the `Single' architecture, we first concatenate the voxel features of two consecutive frames at the feature dimension and then extract the concatenated features using the same encoder as in the main experiments. Such a single-path architecture lacks effective integration of target information from adjacent frame, resulting in a performance gap compared to the `Siamese' counterpart. Conversely, Siamese architecture (denoted as `Siamese') decouples feature extraction from temporal fusion, significantly enhancing tracking performance. For further analysis on the weight-sharing mechanism, we employ an independent dual-stream architecture (denoted as `Dual'), which exhibits a minor decrease in performance. Note that the Siamese architecture enjoys a faster inference speed and fewer parameters compared to the dual-stream architecture. 

\input{Table/bfe}
\paragraph{Effectiveness of BFE} We study the effectiveness of the BFE module in Table~\ref{bfe}. Since the size of rigid objects remains constant during motion, we utilize this prior knowledge in the tracking of rigid objects through BFE. For experiments that disable the BFE, we regress the relative target motion from the motion features directly. The models with BFE significantly outperform the models without BFE on the KITTI and NuScenes for all rigid objects. The results indicate that the BFE can effectively help the tracker to localize the target more accurately. 

\input{Table/scales}
\paragraph{Ablations on STFA} To investigate the impact of multi-scale aggregation in STFA, we train four variants with different aggregation methods, as shown in Table~\ref{scales}. We empirically find that using S3 alone already surpasses the previous SOTA method. Furthermore, combining feature maps from two scales yields additional improvements, except for the Van category. After aggregating feature maps from all three scales, both Success and Precision are steadily improved in all categories. Consequently, we adopt this configuration as the default setting.

\paragraph{Robustness to sparsity} In practical applications, objects captured by LiDAR sensors are predominantly sparse and incomplete. To verify the robustness of SiamMo in sparse scenarios, we divide KITTI and NuScenes into multiple subsets based on different levels of sparsity. Among these, the subset where the number of target points in the first frame falls between [0, 10) represents the most sparse condition, accounting for 37.3\% in the KITTI dataset and 70.7\% in the NuScenes dataset. As shown in Fig.~\ref{fig:sparsity}, owing to the utilization of supplementary information from adjacent frames, SiamMo demonstrates an outstanding advantage over M$^2$-Track in sparse scenarios. Compared to Siamese matching-based methods BAT and PTTR, SiamMo achieves significant 28.9\% and 19.0\% improvements in the NuScenes\_[0, 10) subset. Despite the narrowing performance gap as the number of points approaches 50, SiamMo consistently outperforms all other methods. This indicates that the use of voxelization and BEV representations is effective for the 3D SOT task, and SiamMo performs well in point clouds with varying sparsity. 

\paragraph{Robustness to distractors} For further analysis on the robustness to intra-class distractors, we manually pick out 14 sequences where distractors are around the targets to evaluate the performance of trackers on the KITTI Pedestrian category. As shown in Fig.~\ref{fig:distractors}, apart from SiamMo, other trackers all exhibit significant performance drops. This indicates that SiamMo is robust to distractors, reflecting the potential of SiamMo in being applied to complex real-world situations. 

\paragraph{Efficiency analysis} We report the efficiency of SiamMo in Tab.\ref{efficiency}. It can be observed that SiamMo is lightweight with only 0.82 GFLOPs and 14.6M parameters. Following P2B~\cite{p2b}, we record the average running time of all test frames for the Car category on the KITTI dataset to evaluate the computational efficiency of our method. SiamMo achieves 108 FPS, including 4.2 ms for pre/processing point cloud and 5.0 ms for network forward propagation on a single NVIDIA 4090 GPU. Under the same hardware setup, SiamMo runs faster than the previous SOTA method MBPTrack~\cite{mbptrack} and CXTrack~\cite{cxtrack}.

\begin{figure}[t]
    \centering
    \includegraphics[width=\linewidth]{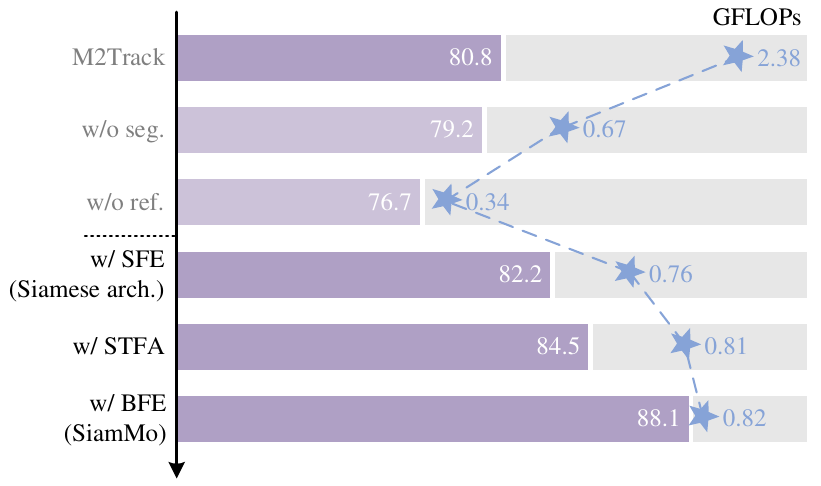}
    \caption{Step-by-step gain of the roadmap.}  
    \label{fig:improvements}
\end{figure}

\input{Table/infer}

Fig.~\ref{fig:improvements} depicts the transformation path from M$^2$-Track to SiamMo, evaluated on the KITTI Car dataset using the Precision metric. 1) We first remove the segmentation module from M$^2$-Track, reducing Precision from 80.8\% to 79.2\% while decreasing the network FLOPs. The single-stream architecture of M$^2$-Track struggles in learning discriminative motion features, resulting in susceptibility to background points. 2) Further removing the refinement module from M$^2$-Track makes a pure motion-centric tracker. This reduces the computational complexity to 0.34G FLOPs, but leads to a significant Precision drop to 76.7\%.  3) Transitioning to a Siamese architecture (with SFE) boosts Precision to 82.2\%, emphasizing the importance of utilizing Siamese features. 4) STFA enhances Precision by capturing diverse motion patterns. 5) Additionally, introducing BFE injects explicit size priors into motion features, resulting in our final SiamMo model with 88.1\% Precision with minor increase in the FLOPs. Compared to the multi-stage motion-centric tracker M$^2$-Track, SiamMo demonstrates advantages in both complexity (from 2.38G to 0.82G) and performance (from 80.8 to 88.1).

\section{Conclusion}
This work presents a novel Siamese motion-centric tracking approach, \textit{i.e.}, SiamMo. It achieves Siamese feature extraction and motion modeling in an end-to-end simple single-stage tracking pipeline. Specifically, we devise a Spatio-Temporal Feature Aggregation (STFA) module that integrates features from adjacent frames at multiple scales to capture motion information effectively. Furthermore, we design a Box-aware Feature Encoding (BFE) that leverages explicit target box priors for motion prediction. Extensive experiments demonstrate that SiamMo is effective and fast, outperforming state-of-the-art methods on three challenging benchmarks, including KITTI, NuScenes, and Waymo Open Dataset, while running at a speed of 108 FPS. Moreover, SiamMo also shows impressive robustness to sparse point clouds as well as distractors. We hope this study could offer valuable insights into 3D SOT research and inspire further exploration of the Siamese motion-centric paradigm.

\bibliographystyle{IEEEtran}
\bibliography{ref}

\vspace{-1cm}
\begin{IEEEbiography}[{\includegraphics[width=1in,height=1.25in,clip,keepaspectratio]{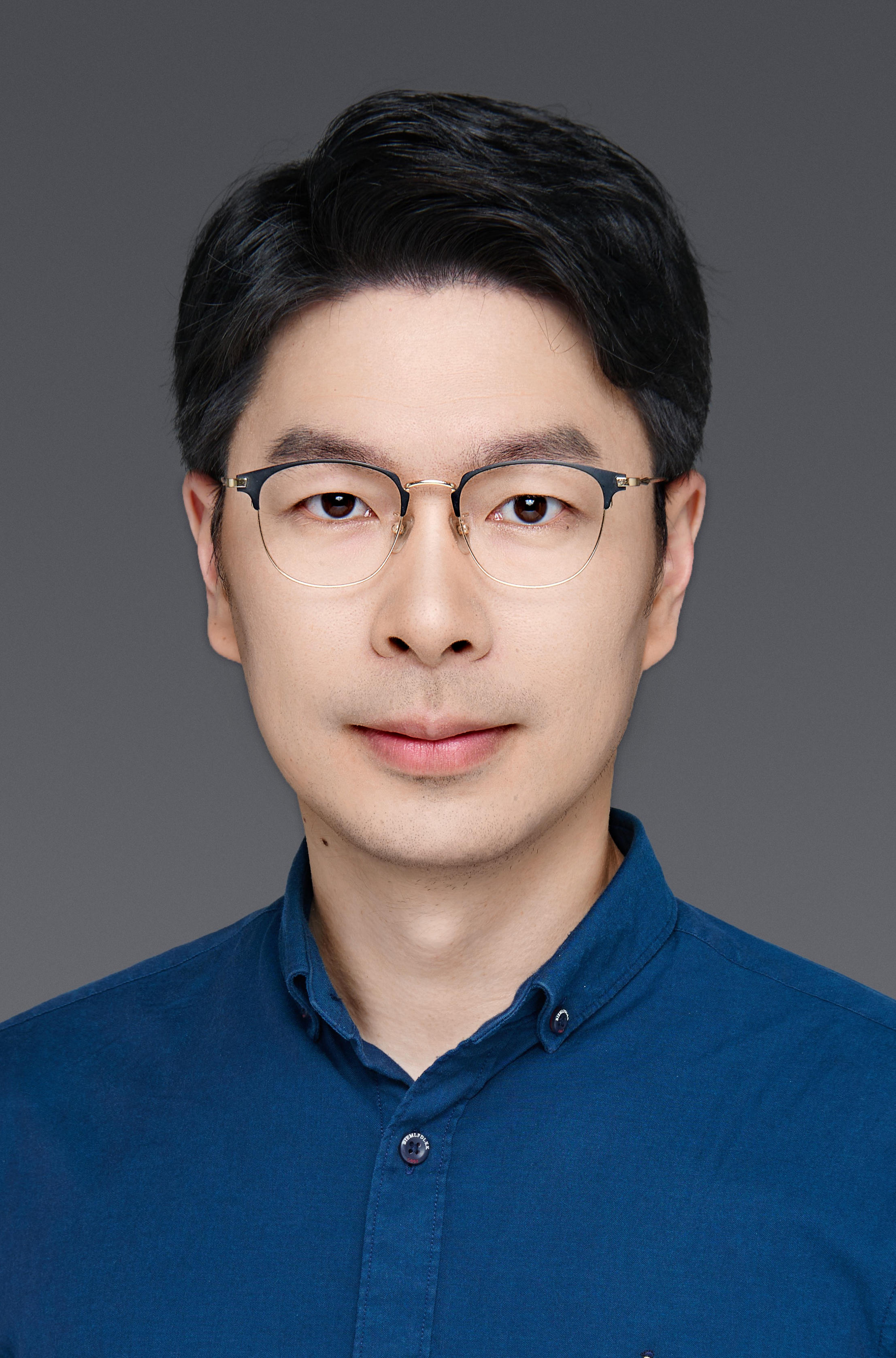}}]
{Yuxiang Yang} received the B.S. and Ph.D. degrees in Control Science and Engineering from University of Science and Technology of China in 2008 and 2013, respectively. He joined Hangzhou Dianzi University, China, in 2013, where he is currently a professor in the School of Electronic and Information. His research interests include computer vision and artificial intelligence. He has published over 30 papers on
important journals and conferences including IJCV, T-MM, NeurIPS, CVPR, IJCAI, AAAI.
\end{IEEEbiography}

\vspace{-1cm}
\begin{IEEEbiography}[{\includegraphics[width=1in,height=1.25in,clip,keepaspectratio]{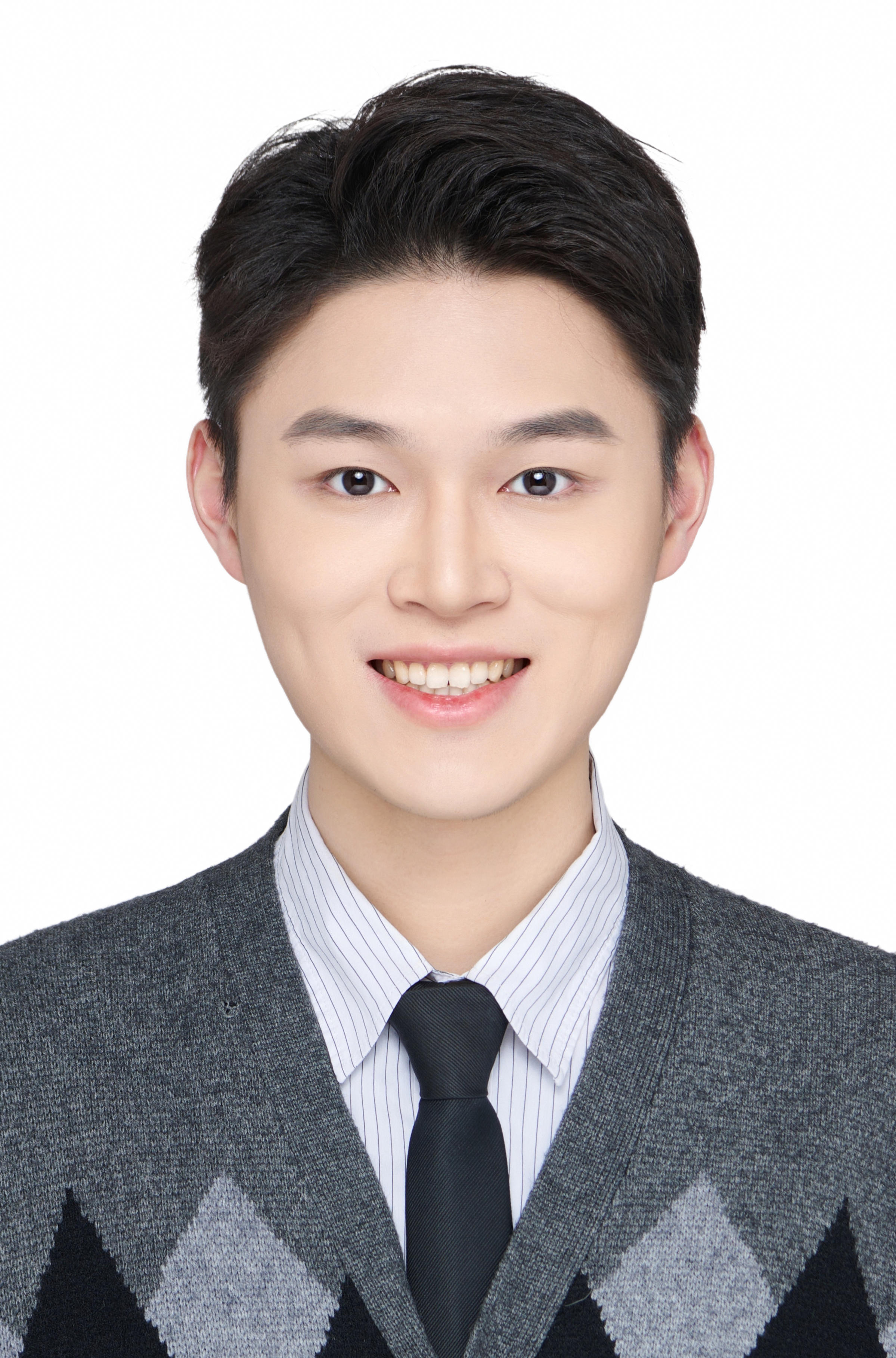}}]
{Yingqi Deng} was born in 2000, and received the B.S. degree in Internet of Things Engineering from Hangzhou Normal University, China, in 2022. Now, he is a M.S. student in the School of Electronic and Information, Hangzhou Dianzi University, China. His research interests include deep learning and autonomous driving. He has participated in the development of multiple autonomous driving projects. He is currently focusing on autonomous driving perception.
\end{IEEEbiography}

\vspace{-1cm}
\begin{IEEEbiography}[{\includegraphics[width=1in,height=1.25in,clip,keepaspectratio]{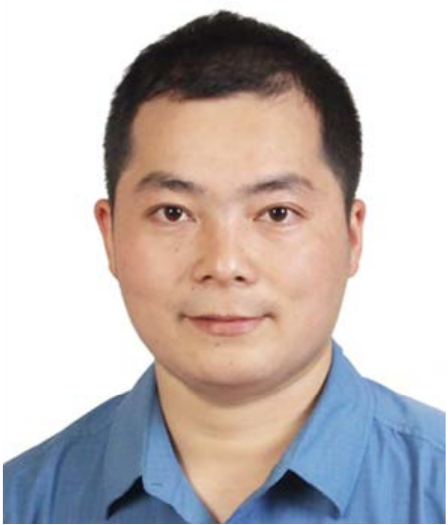}}]
{Jing Zhang} (Senior Member, IEEE) is currently a Research Fellow at the School of Computer Science, The University of Sydney. He has authored over 80 papers in prestigious conferences and journals, including CVPR, ICCV, ECCV, NeurIPS, ICLR, IEEE TPAMI, and IJCV. His research focuses on computer vision and deep learning. Additionally, he is an Area Chair for ICPR, a Senior Program Committee member for AAAI and IJCAI, and a guest editor for IEEE TBD. He regularly reviews for numerous prestigious journals and conferences.
\end{IEEEbiography}

\vspace{-1cm}
\begin{IEEEbiography}[{\includegraphics[width=1in,height=1.25in,clip,keepaspectratio]{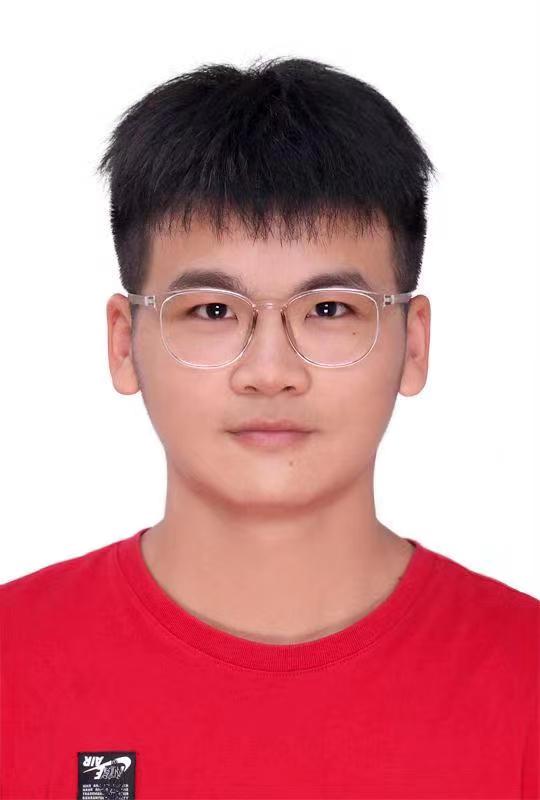}}]
{Hongjie Gu} was born in 2000, and received B.S degree in Material Science and Engineering from China Jiliang University, China, in 2023. Now, he is a M.S student in the School of Electronic and Information, Hangzhou Dianzi University, China. His research interests include computer vision and deep learning. He participated in the development of multiple 3D road perception projects. His current focus is on 3D object tracking.
\end{IEEEbiography}

\vspace{-1cm}
\begin{IEEEbiography}[{\includegraphics[width=1in,height=1.25in,clip,keepaspectratio]{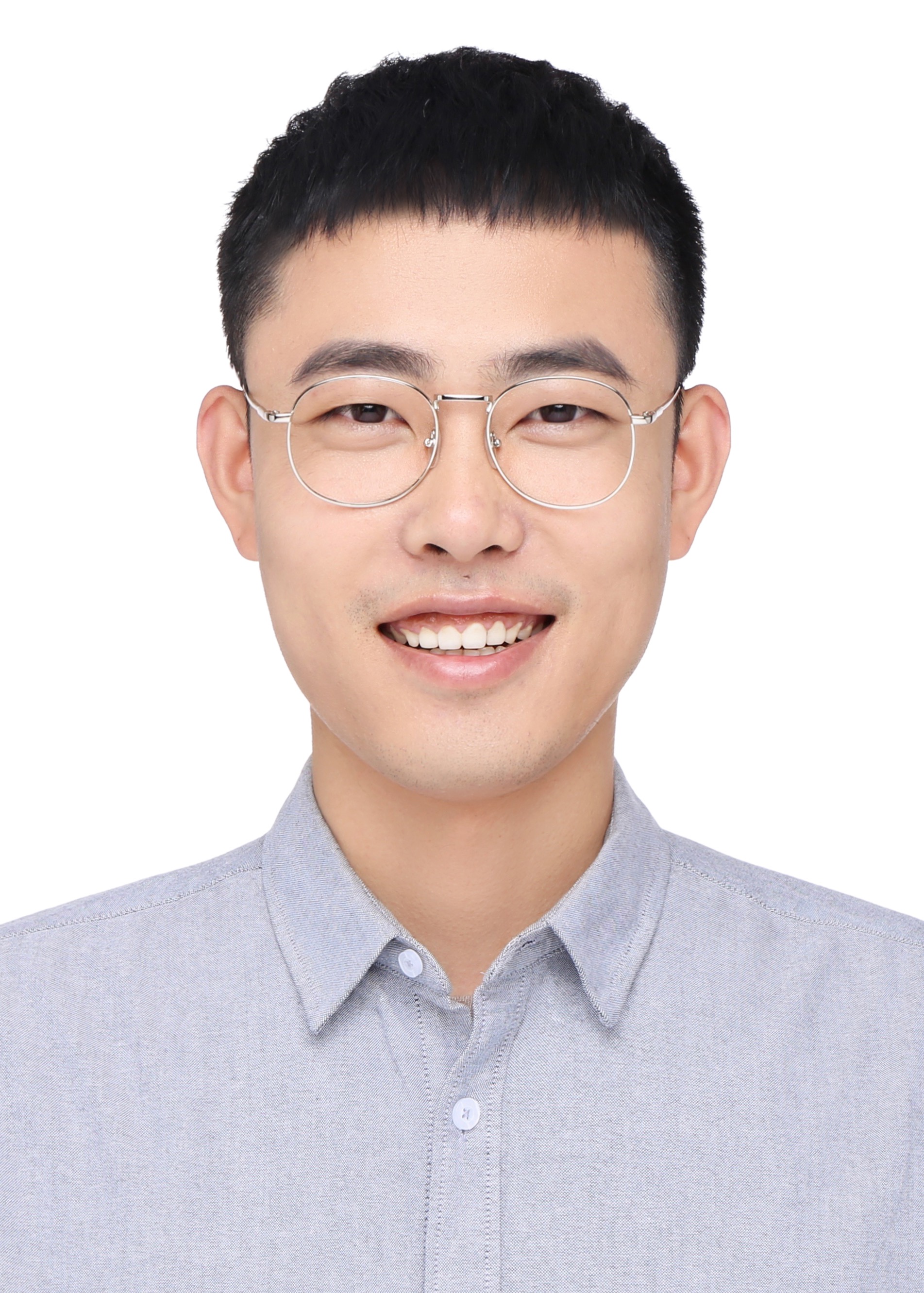}}]
{Zhekang Dong} (Senior Member, IEEE) received the B.E. and M.E. degrees in electronics and information engineering in 2012 and 2015, respectively, from Southwest University, Chongqing, China. He received the Ph.D. degree in control science and engineering from the School of Electrical Engineering, Zhejiang University, China, in 2019. Currently, he is an associate professor in Hangzhou Dianzi University, Hangzhou, China. He is also a Research Assistant (Joint-Supervision) at The Hong Kong Polytechnic University. His research interests cover memristor and memristive system, artificial neural network, the design and analysis of nonlinear systems based on memristor and computer simulation.
\end{IEEEbiography}

\end{document}

%% file: Table/kitti.tex
\begin{table*}[t]
\centering
    \resizebox{0.9\textwidth}{!}{
    \normalsize
    \begin{tabular}{c|c|ccccc}
    \toprule[0.4mm]
\rowcolor{black!10}
Method    & Source  & Car {[}6,424{]} & Pedestrian {[}6,088{]} & Van {[}1,248{]} & Cyclist {[}308{]} & Mean {[}14,068{]} \\
\midrule
\textbf{SiamMo}    & Ours    & \textcolor{red}{\textbf{76.3}} / \textcolor{red}{\textbf{88.1}}    & \textcolor{red}{\textbf{68.6}} / \textcolor{red}{\textbf{93.9}}           & \textcolor{red}{\textbf{67.9}} / \textcolor{red}{\textbf{80.5}}    & \textcolor{red}{\textbf{78.5}} / \textcolor{red}{\textbf{94.8}}      & \textcolor{red}{\textbf{72.3}} / \textcolor{red}{\textbf{90.1}} \\     
M$^2$-Track \cite{m2track}  & CVPR22  & 65.5 / 80.8    & 61.5 / 88.2           & 53.8 / 70.7    & 73.2 / 93.5      & 62.9 / 83.4 \\   
\midrule
MBPTrack \cite{mbptrack} & ICCV23 & 73.4 / 84.8    & \textcolor{red}{\textbf{68.6}} / \textcolor{red}{\textbf{93.9}}           & \textcolor{blue}{\textbf{61.3}} / \textcolor{blue}{\textbf{72.7}}    & \textcolor{blue}{\textbf{76.7}} / \textcolor{blue}{\textbf{94.3}}      & \textcolor{blue}{\textbf{70.3}} / \textcolor{blue}{\textbf{87.9}} \\
SyncTrack \cite{synctrack} & ICCV23 & 73.3 / \textcolor{blue}{\textbf{85.0}}    & 54.7 / 80.5           & 60.3 / 70.0    & 73.1 / 93.8      & 64.1 / 81.9 \\
CXTrack \cite{cxtrack}   & CVPR23 & 69.1 / 81.6    & \textcolor{blue}{\textbf{67.0}} / \textcolor{blue}{\textbf{91.5}}          & 60.0 / 71.8    & 74.2 / 94.3      & 67.5 / 85.3 \\    
CorpNet\cite{corpnet} & CVPRw23 & \textcolor{blue}{\textbf{73.6}} / 84.1 & 55.6 / 82.4 & 58.7 / 66.5 & 74.3 / 94.2 & 64.5 / 82.0 \\
OSP2B \cite{osp2b}     & IJCAI23 & 67.5 / 82.3    & 53.6 / 85.1           & 56.3 / 66.2    & 65.6 / 90.5      & 60.5 / 82.3 \\
GLT-T \cite{glt}     & AAAI23 & 68.2 / 82.1    & 52.4 / 78.8           & 52.6 / 62.9    & 68.9 / 92.1      & 60.1 / 79.3 \\
TAT\cite{tat} & ACCV22 & 72.2 / 83.3 & 57.4 / 84.4 & 58.9 / 69.2 & 74.2 / 93.9 & 64.7 / 82.8 \\
CMT \cite{cmt}       & ECCV22 & 70.5 / 81.9    & 49.1 / 75.5           & 54.1 / 64.1    & 55.1 / 82.4      & 59.4 / 77.6  \\
STNet \cite{stnet}     & ECCV22 & 72.1 / 84.0    & 49.9 / 77.2           & 58.0 / 70.6    & 73.5 / 93.7      & 61.3 / 80.1  \\ 
PTTR \cite{pttr}      & CVPR22 & 65.2 / 77.4    & 50.9 / 81.6           & 52.5 / 61.8    & 65.1 / 90.5      & 57.9 / 78.2 \\
V2B \cite{v2b}       & NIPS21 & 70.5 / 81.3    & 48.3 / 73.5           & 50.1 / 58.0    & 40.8 / 49.7      & 58.4 / 75.2  \\ 
BAT \cite{bat}       & ICCV21 & 60.5 / 77.7    & 42.1 / 70.1           & 52.4 / 67.0    & 33.7 / 45.4      & 51.2 / 72.8 \\  
P2B \cite{p2b}       & CVPR20 & 56.2 / 72.8    & 28.7 / 49.6           & 40.8 / 48.4    & 32.1 / 44.7      & 42.4 / 60.0  \\  
SC3D \cite{sc3d}     & CVPR19 & 41.3 / 57.9    & 18.2 / 37.8           & 40.4 / 47.0    & 41.5 / 70.4      & 31.2 / 48.5 \\
\bottomrule[0.4mm]
\end{tabular}}
\vspace{-5pt}
\caption{Comparison with state-of-the-art methods on KITTI dataset. The best and second-best results are highlighted in \textcolor{red}{\textbf{red}} and \textcolor{blue}{\textbf{blue}}, respectively. Success / Precision are used for evaluation. The top and bottom sections of the table show the \textit{motion-centric trackers} and the \textit{Siamese matching-based trackers}, respectively.} 
\label{kitti}
\end{table*}

%% file: Table/waymo.tex
\begin{table*}[t]
\centering
    \resizebox{\textwidth}{!}{
    \normalsize
    \begin{tabular}{c|cccc|cccc|c}
    \toprule[0.4mm]
    \rowcolor{black!10}
    \rowcolor{black!10} & \multicolumn{4}{c}{Vehicle} & \multicolumn{4}{c}{Pedestrian} & \\
        \rowcolor{black!10}  & Easy& Medium & Hard & Mean & Easy & Medium & Hard & Mean & \\
       \rowcolor{black!10} \multirow{-3}{*}{Method}  & [67,832]&[61,252]&[56,647]&[185,731]& [85,280]&[82,253]&[74,219]&[241,752] & \multirow{-3}{*}{Mean [427,483]}\\
          \midrule
\textbf{SiamMo}                  & \textcolor{red}{\textbf{66.1}} / \textcolor{red}{\textbf{73.4}}  & \textcolor{red}{\textbf{58.1}} / \textcolor{red}{\textbf{67.9}}  & \textcolor{red}{\textbf{56.6}} / \textcolor{red}{\textbf{68.7}}  & \textcolor{red}{\textbf{60.6}} / \textcolor{red}{\textbf{70.1}}   & \textcolor{red}{\textbf{44.8}} / \textcolor{red}{\textbf{66.2}}  & \textcolor{red}{\textbf{35.6}} / \textcolor{red}{\textbf{56.8}}  & \textcolor{red}{\textbf{31.2}} / \textcolor{red}{\textbf{51.8}}  & \textcolor{red}{\textbf{37.5}} / \textcolor{red}{\textbf{58.6}}   & \textcolor{red}{\textbf{47.5}} / \textcolor{red}{\textbf{63.6}}                         \\
\midrule
CXTrack \cite{cxtrack}                 & 63.9 / 71.1  & 54.2 / 62.7  & 52.1 / 63.7  & 57.1 / 66.1   & \textbf{\textcolor{blue}{35.4}} / \textcolor{blue}{\textbf{55.3}}  & \textbf{\textcolor{blue}{29.7}} / \textbf{\textcolor{blue}{47.9}}  & \textbf{\textcolor{blue}{26.3}} / \textbf{\textcolor{blue}{44.4}}  & \textbf{\textcolor{blue}{30.7}} / \textbf{\textcolor{blue}{49.4}}   & \textbf{\textcolor{blue}{42.2}} / \textbf{\textcolor{blue}{56.7}}                        \\
TAT \cite{tat}                     & \textbf{\textcolor{blue}{66.0}} / 72.6  & 56.6 / 64.2  & 52.9 / 62.5  & 58.9 / 66.7   & 32.1 / 49.5  & 25.6 / 40.3  & 21.8 / 35.9  & 26.7 / 42.2   & 40.7 / 52.8                         \\
STNet \cite{stnet}                   & 65.9 / \textbf{\textcolor{blue}{72.7}}  & \textcolor{blue}{\textbf{57.5}} / \textcolor{blue}{\textbf{66.0}}  & \textcolor{blue}{\textbf{54.6}} / \textcolor{blue}{\textbf{64.7}}  & \textcolor{blue}{\textbf{59.7}} / \textcolor{blue}{\textbf{68.0}}   & 29.2 / 45.3  & 24.7 / 38.2  & 22.2 / 35.8  & 25.5 / 39.9   & 40.4 / 52.1                         \\
V2B \cite{v2b}                     & 64.5 / 71.5  & 55.1 / 63.2  & 52.0 / 62.0  & 57.6 / 65.9   & 27.9 / 43.9  & 22.5 / 36.2  & 20.1 / 33.1  & 23.7 / 37.9   & 38.4 / 50.1                         \\
BAT \cite{bat}                    & 61.0 / 68.3  & 53.3 / 60.9  & 48.9 / 57.8  & 54.7 / 62.7   & 19.3 / 32.6  & 17.8 / 29.8  & 17.2 / 28.3  & 18.2 / 30.3   & 34.1 / 44.4                         \\
P2B \cite{p2b}                    & 57.1 / 65.4  & 52.0 / 60.7  & 47.9 / 58.5  & 52.6 / 61.7   & 18.1 / 30.8  & 17.8 / 30.0  & 17.7 / 29.3  & 17.9 / 30.1   & 33.0 / 43.8                        \\
    \bottomrule[0.4mm]
    \end{tabular}}
    \vspace{-5pt}
\caption{Comparison with state-of-the-art methods on Waymo Open Dataset.}
\label{waymo}
\end{table*}

%% file: Table/nuscenes.tex
\begin{table*}[t]
\centering
    \resizebox{\textwidth}{!}{
    \normalsize
    \begin{tabular}{c|ccccc|c|c}
    \toprule[0.4mm]
    \rowcolor{black!10}
    Method   & Car {[}64,159{]} & Pedestrian {[}33,227{]} & Truck {[}13,587{]} & Trailer {[}3,352{]} & Bus {[}2,953{]} & Mean {[}117,278{]} & Mean by Category \\
    \midrule
    \textbf{SiamMo}   & \textcolor{red}{\textbf{64.95}} / \textcolor{red}{\textbf{72.24}}    & \textcolor{red}{\textbf{46.23}} / \textcolor{red}{\textbf{76.25}}           & \textcolor{red}{\textbf{68.22}} / \textcolor{red}{\textbf{68.81}}      &\textcolor{red}{ \textbf{74.21}} / \textcolor{red}{\textbf{70.63}}       & \textcolor{red}{\textbf{65.63}} / \textcolor{red}{\textbf{62.07}}   & \textcolor{red}{\textbf{60.31}} / \textcolor{red}{\textbf{72.68}}      & \textcolor{red}{\textbf{63.85}} / \textcolor{red}{\textbf{70.00}}    \\
    M$^2$-Track \cite{m2track} & 55.85 / 65.09    & 32.10 / 60.92           & 57.36 / 59.54      & 57.61 / 58.26       & 51.39 / 51.44   & 49.23 / 62.73     & 50.86 / 59.05    \\
    \midrule
    MBPTrack \cite{mbptrack}  & \textcolor{blue}{\textbf{62.47}} / \textcolor{blue}{\textbf{70.41}}  & \textcolor{blue}{\textbf{45.32}} / \textcolor{blue}{\textbf{74.03}}  & \textcolor{blue}{\textbf{62.18}} / \textcolor{blue}{\textbf{63.31}}  & \textcolor{blue}{\textbf{65.14}} / \textcolor{blue}{\textbf{61.33}}  & \textcolor{blue}{\textbf{55.41}} / \textcolor{blue}{\textbf{51.76}}  & \textcolor{blue}{\textbf{57.48}} / \textcolor{blue}{\textbf{69.88}}   &  \textcolor{blue}{\textbf{58.10}} / \textcolor{blue}{\textbf{64.17}} \\
    GLT-T \cite{glt}    & 48.52 / 54.29    & 31.74 / 56.49           & 52.74 / 51.43      & 57.60 / 52.01       & 44.55 / 40.69   & 44.42 / 54.33      & 47.03 / 50.98    \\
    PTTR \cite{pttr}     & 51.89 / 58.61    & 29.90 / 45.09           & 45.30 / 44.74      & 45.87 / 38.36       & 43.14 / 37.74   & 44.50 / 52.07      & 43.22 / 44.91    \\
    BAT \cite{bat}      & 40.73 / 43.29    & 28.83 / 53.32           & 45.34 / 42.58      & 52.59 / 44.89       & 35.44 / 28.01   & 38.10 / 45.71      & 40.59 / 42.42    \\
    PTT \cite{ptt}      & 41.22 / 45.26    & 19.33 / 32.03           & 50.23 / 48.56      & 51.70 / 46.50       & 39.40 / 36.70   & 36.33 / 41.72      & 40.38 / 41.81    \\
    P2B \cite{p2b}      & 38.81 / 43.18    & 28.39 / 52.24           & 42.95 / 41.59      & 48.96 / 40.05       & 32.95 / 27.41   & 36.48 / 45.08      & 38.41 / 40.90    \\
    SC3D \cite{sc3d}    & 22.31 / 21.93    & 11.29 / 12.65           & 30.67 / 27.73      & 35.28 / 28.12       & 29.35 / 24.08   & 20.70 / 20.20      & 25.78 / 22.90    \\
    \bottomrule[0.4mm]
\end{tabular}}
\vspace{-5pt}
\caption{Comparison with state-of-the-art methods on NuScenes dataset.}
\vspace{-5pt}
\label{nuscenes}
\end{table*}

%% file: Table/siamese.tex
\begin{table}[t]
\resizebox{\linewidth}{!}{
\begin{tabular}{c|cccc}
\toprule[0.4mm]
\rowcolor{black!10}
Arch. & Car       & Pedestrian & Van       & Cyclist    \\
\midrule
Single    & 72.5 / 84.3 & 61.8 / 88.1  & 61.7 / 74.8 & 72.1 / 93.6 \\
Dual    & 76.0 / 87.8 & 67.5 / 93.1  & 67.2 / 79.7 & 75.1 / 94.2 \\
Siamese & 76.3 / 88.1 & 68.6 / 93.9  & 67.9 / 80.5 & 78.5 / 94.8  \\
\bottomrule[0.4mm]
\end{tabular}}
\caption{Effectiveness of the Siamese architecture.}
\label{siamese}
\end{table}

%% file: Table/bfe.tex
\begin{table}[t]
\resizebox{\linewidth}{!}{
\begin{tabular}{c|cc|cccc}
\toprule[0.4mm]
\rowcolor{black!10} & \multicolumn{2}{c}{KITTI} & \multicolumn{4}{c}{NuScenes}                          \\
\rowcolor{black!10}
\multirow{-2}{*}{BFE} & Car         & Van         & Car         & Truck       & Trailer     & Bus         \\
\midrule
                     & 73.4 / 84.5 & 64.4 / 75.2 & 62.9 / 70.2 & 66.7 / 67.3 & 71.7 / 67.7  & 61.9 / 57.9 \\
\checkmark                     & 76.3 / 88.1 & 67.9 / 80.5 & 65.0 / 72.2 & 68.2 / 68.8 & 74.2 / 70.6 & 65.6 / 62.1 \\
\bottomrule[0.4mm]
\end{tabular}}
\caption{Effectiveness of Box-aware Feature Encoding.}
\label{bfe}
\end{table}

%% file: Table/scales.tex
\begin{table}[t]
\resizebox{\linewidth}{!}{
\begin{tabular}{ccc|cccc}
\toprule[0.4mm]
\rowcolor{black!10}
S1   & S2   & S3   & Car       & Pedestrian & Van       & Cyclist   \\
\midrule
     &      & \checkmark & 74.1 / 85.5 & 66.6 / 90.8  & 67.7 / 80.8 & 73.8 / 94.0 \\
     & \checkmark & \checkmark & 75.5 / 87.5 & 67.1 / 92.0  & 67.5 / 77.5 & 74.2 / 93.7 \\
\checkmark &      & \checkmark & 74.2 / 87.6 & 67.5 / 92.8  & 64.2 / 77.3 & 74.5 / 94.4 \\
\checkmark & \checkmark & \checkmark & 76.3 / 88.1 & 68.6 / 93.9  & 67.9 / 80.5 & 78.5 / 94.8 \\
\bottomrule[0.4mm]
\end{tabular}}
\caption{Ablation studies of STFA. S1, S2, S3 represent multi-scale BEV feature maps at the feature strides of $\{1,2,4\}$, respectively.}
\label{scales}
\end{table}

%% file: Table/infer.tex
\begin{table}[t]
\centering
\resizebox{0.8\linewidth}{!}{
\begin{tabular}{c|ccc}
\toprule[0.4mm]
\rowcolor{black!10}
Method & \#Params       & FLOPs  & Infer speed \\
\midrule
SiamMo    & 14.6M & 0.82G  & 108FPS \\
MBPTrack  & 7.4M & 2.88G  & 74FPS \\
CXTrack & 18.3M & 4.63G  & 48FPS  \\
\bottomrule[0.4mm]
\end{tabular}}
\caption{Efficiency analysis.}
\label{efficiency}
\end{table}